\documentclass[runningheads]{llncs}

\usepackage{eccv}

\usepackage{eccvabbrv}

\usepackage{graphicx}
\usepackage{booktabs}

\usepackage[accsupp]{axessibility}  % Improves PDF readability for those with disabilities.

\usepackage{hyperref}

\usepackage{orcidlink}

\usepackage{listings}
\usepackage[table]{xcolor}

\begin{document}

\title{Think--Verify--Revise: Neuro-Symbolic Visual Reasoning with Vision-Language Models and Dynamic Logic Tensor Networks} 

%\titlerunning{D-LTN for NeSy Visual Reasoning}
\titlerunning{NeSy Visual Reasoning with VLM and D-LTN}

% Second Author\inst{2,3}\orcidlink{1111-2222-3333-4444} \and
% Third Author\inst{3}\orcidlink{2222--3333-4444-5555}}
\author{Homayoun Afshari \and
Pietro Basci \orcidlink{0009-0008-3335-6675} \and
Alessandro Russo \orcidlink{0009-0009-6711-8277} \and
Lia Morra \orcidlink{0000-0003-2122-7178}}

\authorrunning{H.~Afshari et al.}

% Springer Heidelberg, Tiergartenstr.~17, 69121 Heidelberg, Germany
\institute{Politecnico di Torino \\Department of Control and Computer Engineering \\Turin, Italy\\
\email{\{homayoun.afshari\}@studenti.polito.it}\\
\email{\{pietro.basci, alessandro.russo, lia.morra\}@polito.it}}

\maketitle

\begin{abstract}
Visual reasoning tasks require a system to jointly perceive visual content and apply formal relational constraints---a combination that neither pure neural nor purely symbolic approaches handle well in isolation. This paper proposes a Neuro-Symbolic (NeSy) framework that closes this gap by tightly coupling a Vision-Language Model (VLM) for \emph{automatic First-Order Logic (FOL) rule induction} with a Dynamic Logic Tensor Network (D-LTN) for \emph{differentiable rule verification}, in a closed iterative feedback loop. The VLM receives a small set of labelled visual examples and proposes candidate FOL rules conforming to a strict grammar (Think); the D-LTN is automatically assembled from these rules at runtime and evaluates them grounding on CNN-produced visual embeddings (Verify); and verification failures are fed back to guide the VLM's next hypothesis (Revise). Evaluated on the ViSudo-PC benchmark across four visual domains (MNIST, EMNIST, KMNIST, FMNIST), the system induces valid Sudoku constraint rules using only three training examples as visual context. The proposed method achieves AUC scores matching or outperforming previous methods (NeuPSL, LTN), showing the potential for automatic rule discovery through VLM. Code is available at \url{https://github.com/homayoun-afshari/nesy}.


\end{abstract}

\begin{figure}[ht]
    \centering
    \includegraphics[width=0.9\linewidth]{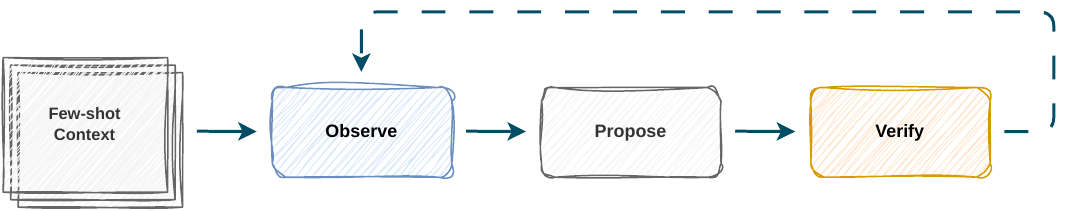}
    \caption{Conceptual diagram of the proposed feedback-driven neuro-symbolic reasoning framework. Given a few labelled visual examples, the system generates candidate symbolic rules, verifies them against visual evidence, and uses verification outcomes as feedback to iteratively refine its hypotheses. This closed-loop process enables the automatic discovery of valid symbolic constraints from visual data.}
    \label{fig:conceptual}
\end{figure}

% ============================================================
\section{Introduction}
\label{sec:intro}
% ============================================================

Visual reasoning tasks---such as determining whether a visually encoded Sudoku board is valid, or identifying the abstract rule governing a sequence of images---require a system to jointly perceive visual content and apply formal relational constraints. Pure neural approaches struggle with the latter: even powerful Vision-Language Models (VLMs) that excel at inductive pattern recognition fail to reliably chain logical rules or generalise to unseen constraint combinations~\cite{zhang2024nesy,hamilton2024}.
Conversely, traditional Neuro-Symbolic (NeSy) systems that embed rules as hard-coded constraints achieve transparency but, with exceptions, they require a human expert to hand-craft the correct logical knowledge for every new task~\cite{morra2023ltn,daniele2024transfer,badreddine2022ltn}. To achieve systems that are simultaneously flexible, explainable, and based on formal language, a central challenge to tackle is  therefore \emph{where rules come from}: can they be induced automatically from data in a form that is both machine-verifiable and generalisable across visual domains?

This paper proposes a NeSy visual reasoning framework that closes the above gap by tightly coupling a VLM-based rule inducer with a Dynamic Logic Tensor Network (D-LTN) verifier in a closed feedback loop (Figure~\ref{fig:conceptual}). The key idea is to use the VLM not merely as a static rule writer, but as an abductive reasoner whose hypotheses are continuously evaluated and corrected by a differentiable symbolic verifier. Concretely, a \textbf{Rule Generator (VLM)} receives a textual description of the task and a small set of visual examples (context) and proposes candidate rules in a formally defined FOL grammar. 
A \textbf{Rule Verifier (D-LTN)} automatically parses the FOL rule, constructs the corresponding LTN computation graph at runtime, grounding the symbolic layers on the embedding computed by a \textbf{Visual Encoder (CNN)}. The D-LTN is then trained on the target task and the satisfiability of the extracted rules serves as a feedback signal to the Rule Generator. 
Owing to the differentiable nature of LTNs, gradients can be propagated through the verifier, enabling end-to-end training of the visual encoder. Our approach, while demonstrated on LTNs, could be easily extended to other NeSy frameworks, e.g., NeuPSL \cite{pryor2022neupsl}, based on comparable logic languages. 
Finally, the \textbf{Feedback Loop} returns both the verified rule's performance  and any parsing or logical errors to the VLM, which uses this information to refine its next hypothesis. 

We evaluate the framework on the ViSudo-PC benchmark~\cite{augustine2022visudo}, requiring systems to verify visually encoded $4\!\times\!4$ Sudoku boards under four visual domains (MNIST, EMNIST, KMNIST, FMNIST). Our main contributions are as follows:

\begin{itemize}
  \item We introduce a closed-loop framework for \emph{automatic induction of FOL rules from visual data}, combining VLM-based abductive hypothesis generation with D-LTN-based differentiable verification.
  \item We propose a \emph{grammar-constrained prompting strategy} that steers VLM outputs towards syntactically valid and formally parseable FOL rules, reducing hallucinations and ensuring downstream compatibility with the D-LTN verifier.
  \item We demonstrate that the proposed framework can successfully induce valid Sudoku constraint rules using only three training examples as context and a limited number of VLM interactions (3--19 iterations).
  \item We show that, when evaluated independently of the VLM induction component, the proposed CNN+D-LTN architecture \emph{matches or outperforms existing state-of-the-art approaches}, including NeuPSL~\cite{pryor2022neupsl} and LTN-IND A/B/C~\cite{morra2023ltn}, across all four visual domains.
\end{itemize}

% ============================================================
\section{Related Work}
\label{sec:related}
% ============================================================

\subsection{Visual reasoning in NeSy AI}

Visual reasoning serves as a compelling testbed for NeSy architectures, requiring integration of bottom-up perception with top-down logical inference to interpret scenes, detect causal relations, and apply structured reasoning over visual inputs~\cite{liang2022visual}.
Key benchmarks include \emph{Visual Abductive Reasoning} (VAR)~\cite{liang2022visual}, \emph{VideoABC}~\cite{zhao2022videoabc}, \emph{Raven's Progressive Matrices} (RPM) and its variants~\cite{camposampiero2024arlc}, \emph{CLEVR\allowbreak}~\cite{santoro2017relational} for relational VQA, and \emph{ViSudo-PC}~\cite{augustine2022visudo}.
ViSudo-PC blends visual perception with symbolic relational constraints, requiring systems to determine whether a visually rendered Sudoku grid is correctly solved from images drawn from MNIST, EMNIST, KMNIST, and FMNIST. It serves as the primary testbed in this paper.

\subsection{LLM-Based Rule Induction for NeSy Systems}

A growing body of work combines neural perception with symbolic rule learning through Large Language Models (LLMs) or AI agents. \emph{Concept-RuleNet}~\cite{conceptrulenet} is a multi-agent neurosymbolic system in which a multimodal generator mines discriminative visual concepts directly from training images, an LLM reasoner composes them into first-order rules, and a VLM verifier scores symbol presence at inference; its agents are frozen, prompted models rather than jointly trained components. \emph{ILP-CoT}~\cite{ilpcot} bridges inductive logic programming with multimodal LLMs, using an MLLM to propose structurally-correct rule skeletons that prune the ILP search space before a discrete solver induces rules over rectified facts.
The VLM+ASP system of \cite{vlm_nesy_commonsense} couples a VLM front-end with Answer Set Programming for commonsense visual scene interpretation, exploiting ASP's expressive non-monotonic reasoning. Finally, the \emph{IDEA} framework~\cite{idea2024} couples induction, deduction, and abduction in an LLM-agent loop that abduces hypotheses, deduces plans to test them, and induces refined rules from interactive feedback, structurally echoing our closed feedback loop. Across all of these, the symbolic layer is realised by discrete solvers or frozen, prompted LLM/VLM agents that admit no gradient signal; in contrast, our D-LTN verifier is differentiable and propagates gradients through the symbolic layer back into the visual encoder, enabling end-to-end optimisation of perception against the induced rules.

\begin{figure}[t]
    \centering
    \includegraphics[width=0.9\linewidth]{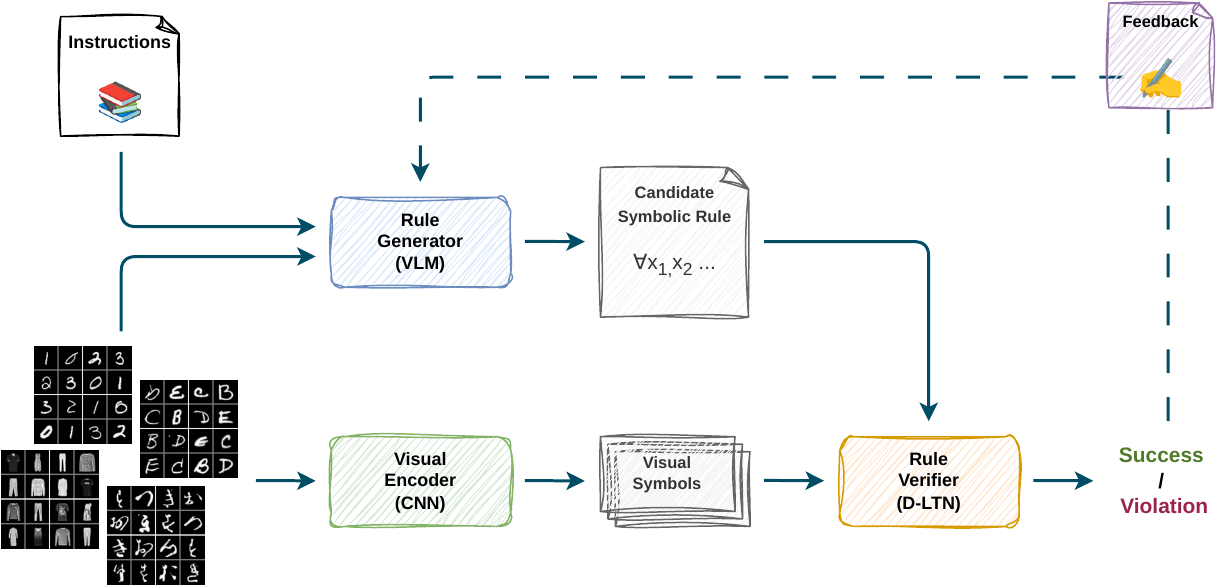}
    \caption{Overview of the proposed neuro-symbolic framework for visual rule discovery. From only three labeled visual Sudoku examples, a \textcolor{blue!50}{Vision-Language Model (VLM)} induces candidate \textcolor{gray}{First-Order Logic (FOL) rules}, which are verified by a dynamically constructed \textcolor{orange!50}{Logic Tensor Network (D-LTN)} operating on CNN-derived \textcolor{gray}{Visual Symbols}. Verification failures are \textcolor{magenta!70}{fed back} to the VLM in a closed-loop process, enabling automatic discovery of valid Sudoku constraints across multiple visual domains.}
    \label{fig:architecture}
\end{figure}

% ============================================================
\section{Methodology}
\label{sec:method}
% ============================================================

In this section, we present the proposed neuro-symbolic framework for automatic rule discovery from visual data. The framework combines neural perception, symbolic rule induction, and differentiable logical reasoning within an iterative feedback loop. Starting from a small set of labeled visual examples, candidate symbolic rules are generated, verified against visual evidence, and progressively refined until valid constraints are discovered. The following subsections describe the overall architecture and its individual components.

\subsection{Framework overview}

Figure~\ref{fig:architecture} illustrates the proposed framework. The architecture consists of four interconnected components: (i) a Rule Generator (VLM), which induces candidate logical rules from a small set of labeled examples; (ii) a Visual Encoder (CNN), which extracts neural representations from image data; (iii) a Rule Verifier (D-LTN), which evaluates the generated rules against the visual representations through differentiable logical reasoning; and (iv) a Feedback Loop, which uses verification outcomes to guide subsequent rule generation. 
The framework operates as a closed iterative process. Candidate FOL rules produced by the Rule Generator are first translated into a symbolic representation and validated through syntactic parsing. The resulting formulas are then used to dynamically construct an LTN, whose terms and predicates are grounded using the visual representations extracted by the Visual Encoder. The Rule Verifier evaluates the degree to which the generated rules are satisfied by the visual evidence, and the resulting feedback is returned to the Rule Generator to refine subsequent hypotheses. This process is repeated until valid symbolic constraints emerge.

\subsection{Rule Generation (Think)}

The Rule Generator module transforms multimodal inputs into symbolic representations. To enable downstream logic-based reasoning, symbol discovery must be combined with rule induction yielding: (i) a list of symbols (functions, predicates, constants); (ii) First-Order Logic (FOL) statements representing the symbolic rules, and (iii) groundings defining the semantics of logical symbols, including constants, predicates, and functions. In principle, an LLM/VLM (or combination of specialized agents) should be flexible enough to produce all of the above as structured outputs. However, to reduce the search space, in this proof-of-concept we fix the available predicates and their Python groundings through prompt engineering, and focus exclusively on outputting the FOL statements.

\paragraph{Prompt engineering.} Textual instructions, visual examples, and intermediate feedback are first combined to construct a coherent context using a combination of few-shot and Chain-of-Thought (CoT) prompting. The resulting prompt is provided to the VLM, which acts as the rule generator by leveraging its multimodal reasoning capabilities to induce symbolic constraints from the input examples. To enable downstream logic-based reasoning, the VLM is instructed to produce FOL rules.  
Additional details on the experimental configuration are provided in Section~\ref{ssec:impl_details}.

\paragraph{Response validation.} Since the raw VLM response may contain intermediate CoT reasoning traces, a post-processing stage applies predefined patterns and Regular Expressions (RegEx) to extract the formal symbolic content while discarding auxiliary reasoning text. The resulting rules and groundings are then forwarded to the verification module.

\subsection{Rule Verification (Verify)}

The Rule Verification module verifies the VLM-generated FOL rule by grounding their predicates in the visual symbols extracted by the CNN.

\paragraph{LTN Fomula generation.} With the decoded FOL rules and Python groundings, we convert the symbolic rules into a D-LTN. A \emph{Rule Parser} converts the FOL rules, stated in a pre-defined grammar, into a \emph{Syntax Tree}. An \emph{LTN Builder} then converts the tree nodes into customized node classes to generate the final \emph{LTN Formula}. At its core, an LTN represents a fuzzified grounding of an FOL rule~\cite{badreddine2022ltn}. The structure of the LTN-based syntax tree is the computational graph of the LTN itself---traversing it is equivalent to feeding inputs into the LTN to compute its truth value. 

\paragraph{Visual Encoding}

The Visual Encoder is responsible for extracting the visual symbols used for grounding in the LTN. It operates independently of the VLM and is used exclusively during the symbolic verification phase. The visual input consists of a batch of images, each containing multiple visual objects. Assuming a batch size of (B), a maximum of (O) objects per image, object dimensions of ($W \times H$), and (C) color channels, the input tensor has shape ($B \times O \times C \times W \times H$). The CNN maps each object to an embedding of dimension (E), producing an output tensor of shape ($B \times O \times E$). The perceptor can be used in a modular (frozen) setting, if pre-trained, or an end-to-end (trainable) setting; in all our experiments, we train the CNN end-to-end. To incorporate positional information, row and column indices are concatenated with the semantic embeddings.

\subsection{Feedback Handling (Revise)}

Once the D-LTN produces its final output, the system generates intermediate feedback consisting of the most recently derived symbolic rules and their performance evaluation. Rules that fail syntactic validation are also included in the feedback, together with the corresponding parsing errors, even though they are not evaluated by the D-LTN. This feedback is incorporated into the next prompt, enabling iterative refinement. A D-LTN is defined as \emph{trainable} if it contains at least one node grounded by a neural network. Given a fuzzified FOL rule producing a truth value $t \in [0,1]$ and a decision threshold $\tau \in [0,1]$, where $\tau$ is a user-defined hyperparameter, the system considers the rule satisfied if $t \geq \tau$. If the threshold is not reached (i.e., $t < \tau$), the generated feedback is used to construct the next prompt, and the refinement process continues iteratively until the rule satisfies the threshold or the maximum number of iterations is reached. Additional details on prompt are provided in Section~\ref{ssec:impl_details}.

% ============================================================
\section{Experimental Setup}
\label{sec:setup}
% ============================================================

\subsection{Benchmark and Dataset}

We evaluate our approach on the ViSudo-PC benchmark~\cite{augustine2022visudo}, which focuses on verifying the correctness of visually encoded Sudoku boards. The task consists of determining whether a visually encoded Sudoku board satisfies the Sudoku constraints, requiring both perceptual recognition of the cell contents and logical reasoning over their spatial relationships. Each Sudoku board is represented as a grid of images rather than symbolic labels, making the benchmark particularly suitable for evaluating integrated perception and reasoning systems. The benchmark comprises 11 data splits: the first 10 are used for evaluation, while the 11th is reserved for experimentation. Each split contains training, validation, and test subsets, each consisting of 100 board pairs. Every pair includes one valid and one corrupted 4 × 4 Sudoku board. 
Individual board cells are represented as 28 × 28 grayscale images from four domains of increasing visual complexity: MNIST (digits), EMNIST (English letters), KMNIST (Japanese characters), and FMNIST (fashion items).

\subsection{Baseline methods}

To assess the effectiveness of the proposed framework, we compare against representative neuro-symbolic approaches previously evaluated on the ViSudo-PC benchmark. Specifically, we consider NeuPSL~\cite{pryor2022neupsl} and the three LTN-based variants introduced by~\cite{morra2023ltn}, namely LTN-IND A, LTN-IND B, and LTN-IND C.

For all baselines, we report the results published in the original works and compare them against the performance achieved by our method under the same benchmark setting.

\subsection{Evaluation Metrics}

We evaluate model performance using the Area Under the Receiver Operating Characteristic Curve (AUC). AUC is adopted as the primary evaluation metric, as it provides a threshold-independent measure of the model's ability to distinguish between valid and corrupted Sudoku boards. 
To ensure comparability with prior work, results are aggregated following the evaluation protocol of~\cite{augustine2022visudo}. Specifically, performance is computed independently on each of the ten evaluation splits of the ViSudo-PC benchmark, and the final score is reported as the average AUC across splits. %Accuracy is aggregated in the same manner.
In addition to predictive performance, we assess the correctness of the generated symbolic knowledge. Since multiple first-order logic formulations may express the same Sudoku constraint, the generated rules are not evaluated solely based on syntactic similarity to a reference rule. Instead, their \emph{logical equivalence} is established through formal mathematical analysis, demonstrating that the generated formulations preserve the intended semantics of the original constraints.

\subsection{Implementation details}
\label{ssec:impl_details}

\paragraph{Architecture.} 
The \emph{Rule Generator} is built on \texttt{Llama-4-Maverick-17B-128E-Ins-\\truct} model~\cite{meta2025llama}, accessed via Groq API~\cite{groq2025}. The model employs a Mixture-of-Experts (MoE) architecture with 128 experts and a context window of up to one million tokens. The \emph{Visual Encoder} consists of a CNN composed of stacked convolutional blocks (Conv2D, ReLU, GroupNorm, MaxPool, Dropout). The resulting feature representations are projected through a fully connected layer to produce embeddings of dimension \texttt{embed\_dims}. An optional softmax layer is applied at the output stage. To encode spatial information, row and column indices are concatenated with the visual embeddings before they are processed by the reasoning module. For the \emph{Rule Verifier}, we adopt the same grounding as~\cite{morra2023ltn}. Structural predicates ($P_{\text{same\_row}}$, $P_{\text{same\_col}}$, $P_{\text{same\_block}}$, $P_{\text{same\_loc}}$) use binary similarity; the perceptual predicate $P_{\text{same\_value}}$ uses exponential similarity. Logical connectives are grounded by Goguen's product t-norm (AND), Goguen's t-conorm (OR), complement (NOT), Reichenbach's implication (IMPLIES), and linear similarity (IFF), generalized mean w.r.t. the error (FORALL aggregator) and generalized mean (EXISTS aggregator) \cite{badreddine2022ltn,morra2023ltn}. 
During rule induction, the Think–Verify–Revise loop terminates when the corresponding D-LTN achieves an $AUC \geq 0.95$ on the test set, at which point the candidate rule is considered valid. The threshold was chosen empirically to balance rule quality and the number of refinement iterations. Otherwise, the verification feedback is incorporated into the next prompt to guide rule refinement.

\paragraph{Prompt.} 
To improve the reliability of rule generation, we constrained the VLM output space through a predefined vocabulary of pre-grounded predicates. The model was therefore tasked solely with generating symbolic FOL rules describing relationships among visual objects and was restricted to producing a single rule per iteration.
To support iterative refinement, the prompt incorporates a history of previous iterations, including the generated rules and their corresponding performance scores, enabling the model to leverage prior outcomes when proposing new rule hypotheses. 
Listing~\ref{list:prompt2} reports the prompt used for rule generation.

\begin{lstlisting}[basicstyle=\scriptsize, caption={VLM prompt used for FOL rule generation. The prompt constrains the search space through a predefined set of grounding alternatives and is iteratively augmented with previously generated rules and their verification scores to guide rule refinement.}, label={list:prompt2}]
    
    system_role = '''
        You are a helpful assistant that can extract the First-Order
        Logic (FOL) rule from images.
        THE GRAMMAR OF FOL:
        - Constants: Not allowed in the rule.
        - Variables: Your options are `x1`, `x2`, ..., which
          represent visual objects.
        - Functions: Not allowed in the rule.
        - Predicates: Your options are `P_same_row`, `P_same_col`,
          `P_same_block`, `P_same_loc`, and `P_same_value`.
        - To compare variables, only use predicates.
        - The symbols used for logical AND, OR, and NOT are
          respectively `&`, `|`, and `!`.
        - The symbols used for implication and equivalence are
          respectively `implies` and `iff`.
        - The symbols used for universal and existential quantifiers
          are respectively `forall` and `exists`.
        - Use parentheses for preserving operation precedence.
        WHAT YOU MUST CONSIDER:
        - Use your own knowledge to analyze and deeply think about the
          images provided as your reference.
        - All the images must follow the same rule that you extract.
        - The rule applies to the visual objects within each image.
        - The visual objects may represent numbers rather than what
          they really are.
        - At the end of your chain of thought, put the extracted rule
          in the following template:
          EXTRACTED_RULE: "the rule you extracted"
    '''
    if len(history_list) > 0:
        n_extracted_rules = 0
        system_role += 'HISTORY OF PREVIOUS TRIALS:'
        for trial, incident in enumerate(history_list):
            error_message, extracted_fol_rule, ratio = incident
            system_role += (
                f'-  Trial {trial+1} -> '
            )
            if error_message != '':
                system_role += (
                    f'error: "{error_message}"'
                )
            else:
                n_extracted_rules += 1
                system_role += (
                    f'extracted rule: "{extracted_fol_rule}", '
                    f'conforming images: {100 * ratio:.2f}%'
                )
        if ratio < termination_threshold:
            system_role += 'IMPORTANT LESSON FROM HISTORY:'
            if n_extracted_rules == 0:
                system_role += (
                    '-  Pay attention to the the instructions!'
                )
            else:
                system_role += (
                    '-  The next FOL rule must be an improved version'
                    ' of the above!'
                )

    prompt = [{
        'type': 'text',
        'text': 'These are the reference images:'
    }]
    for base64_image in base64_image_list:
        prompt.append({
            'type': 'image_url',
            'image_url': {
                'url': f'data:image/png;base64,{base64_image}'
            }
        })
        
    chat_completion = client.chat.completions.create(
        messages=[
            {'role': 'system', 'content': system_role},
            {'role': 'user', 'content': prompt}
        ]
    )

    response = chat_completion.choices[0].message.content
\end{lstlisting}

\paragraph{Training.} The Visual Encoder is trained end-to-end together with the Rule Verifier. Due to the differentiable nature of LTNs, gradients can be propagated through the logical constraints and back to the CNN parameters. 
The training objective is defined as: 

\begin{equation}
  l(x, y) = \begin{cases} f(x) & \text{if } y = 0 \\ 1 - f(x) & \text{if } y = 1, \end{cases}
  \label{eq:loss}
\end{equation}

where $y \in \{0,1\}$ denotes the board label (valid or invalid), and $f$ denotes the composed CNN---D-LTN model. This objective encourages the model to maximize the satisfaction of the encoded logical constraints, in other words, the rule should be satisfied if the board is valid, and not satisfied if the rule is not valid. In multi-rule scenarios, which we leave to future work, the above can be substituted by the satisfiability of the knowledge base.

\paragraph{Hyperparameter Tuning.} Since the quality of the perceptual groundings directly affects the performance of the symbolic verification module, we conducted a hyperparameter search to identify an effective visual encoder configuration. The search space included the number of convolutional layers, kernel size, embedding dimensionality, dropout rate, and the use of a softmax output layer. We searched several CNN configurations across four data sources, on the 11th split using an initial hand-crafted FOL rule: 

\vspace{0.1cm}
\begin{lstlisting}
    forall x1, x2
        P_same_value(x1, x2) implies (
            P_same_loc(x1, x2) | (!P_same_row(x1, x2) & 
                                  !P_same_col(x1, x2) & 
                                  !P_same_block(x1, x2)))
\end{lstlisting}
\vspace{0.1cm}

This rule was introduced as a practical convenience for the current proof-of-concept evaluation and was used exclusively during hyperparameter tuning; it is not an inherent requirement of the rule induction pipeline. The rule was not used in any subsequent experiments. The best CNN configuration was selected according to the highest average AUC and subsequently used for experiments on the evaluation splits. A sufficiently strong visual encoder, or ideally a pretrained perceptual model, could reduce the need for such task-specific tuning and further improve the automation of the overall rule induction process.

% ============================================================
\section{Results}
\label{sec:results}
% ============================================================

\subsection{Visual encoder selection}

The selected configuration, consisting of \texttt{cnn\_dims=(32,64)}, \texttt{kernel\_dims=(4,4)}, \\ \texttt{embed\_dims=(64,)}, \texttt{drop\_prob=0.2}, and \texttt{use\_softmax=True} achieved the best overall performance. This model obtained an average test AUC of 0.9560 and an accuracy of 85.50\%, providing reliable visual representations for the grounding predicates used by the Rule Verifier. The selected encoder was therefore adopted in all subsequent experiments.

\subsection{Symbolic Rule Discovery}

\begin{table}[t]
\centering
\small
\setlength{\tabcolsep}{6pt}
\caption{Test AUC obtained on the ViSudo-PC (11th split) using the FOL rules automatically discovered by the VLM across the MNIST, EMNIST, and KMNIST domains. In the FMNIST domain, the framework did not discover a valid rule before reaching the maximum iteration limit of 20.}
\label{tab:rulegen}
\begin{tabular}{lcccc}
\toprule
\textbf{Data Source} & \textbf{VLM Load} & \textbf{Rule} & \textbf{Iterations} & \textbf{AUC} \\
\midrule
MNIST-11  & 3 & Rule~1 & 19 & 0.9974 \\
EMNIST-11 & 3 & Rule~2 & 13 & 0.9012 \\
KMNIST-11 & 3 & Rule~3 &  9 & 0.9517 \\
FMNIST-11 & 3 & Error  & 20 & --- \\
\bottomrule
\end{tabular}
\end{table}

\begin{figure}[t]
\label{fig:rules}
\begin{lstlisting}[caption={Valid symbolic rules discovered by the VLM for the MNIST, EMNIST, and KMNIST domains. Although the three generated FOL formulations differ syntactically, they are semantically equivalent, indicating that the proposed framework consistently converges to valid representations of the underlying constraint.}, label={list:rules}]

RULE 1 (MNIST):

    forall x1, x2
        !P_same_loc(x1, x2) & (P_same_row(x1, x2) |
                               P_same_col(x1, x2) |
                               P_same_block(x1, x2))
        implies !P_same_value(x1, x2)

RULE 2 (EMNIST):

    forall x1 forall x2
        (P_same_row(x1, x2) |
         P_same_col(x1, x2) |
         P_same_block(x1, x2)) & !P_same_loc(x1, x2)
        implies !P_same_value(x1, x2)

RULE 3 (KMNIST):

    forall x1, x2
        !P_same_loc(x1, x2) & P_same_value(x1, x2) implies (
            !P_same_row(x1, x2) &
            !P_same_col(x1, x2) &
            !P_same_block(x1, x2))    
            
\end{lstlisting}
\end{figure}

Listing~\ref{list:rules} presents the FOL rules automatically generated by the proposed framework for the MNIST, EMNIST, and KMNIST domains. Although the resulting formulas differ syntactically, they are logically equivalent and capture the same underlying Sudoku constraint. In particular, Rules 1 and 2 correspond to the same formula, differing only in the ordering of conjuncts within the antecedent. Since conjunction is commutative, both formulations are logically identical. Rule 3 expresses the same constraint through an alternative formulation. By applying the contrapositive together with De Morgan's law, Rule 3 can be shown to be equivalent to Rules 1 and 2. Consequently, all three generated rules encode the same knowledge: \emph{two distinct cells that belong to the same row, column, or block cannot contain the same value}. The formulas are not only logically equivalent but, under the chosen grounding, they also have equivalent interpretation. The emergence of equivalent formulations across different visual domains indicates that the proposed framework consistently converges to valid symbolic representations of the target constraint despite variations in syntactic structure.
In contrast, no equivalent rule was recovered for FMNIST within the iteration limit of 20. This failure likely reflects a greater difficulty in associating symbolic meaning to clothing items, making it harder for the VLM to extract consistent relational patterns.
The number of VLM iterations required and the corresponding relative AUC achieved on each domain are reported in Table~\ref{tab:rulegen}.

\subsection{Visual Reasoning Performance}

To evaluate the effectiveness of the proposed framework, we selected the first extracted rule and applied it across all ten evaluation splits and the four visual domains of the ViSudo-PC benchmark. Since all extracted rules were shown to be logically equivalent, the choice of a specific rule does not affect the semantics of the reasoning process.

\begin{table}[t]
\setlength{\tabcolsep}{6pt}
\centering
\small
\caption{Comparison with state-of-the-art methods on ViSudo-PC ($4\times4$), reported as test AUC $\pm$ std., averaged over the first 10 splits. The best result for each dataset is shown in \textbf{bold}.}
\label{tab:comparison}
\begin{tabular}{lcccc}
\toprule
\textbf{Method} & \textbf{MNIST} & \textbf{EMNIST} & \textbf{KMNIST} & \textbf{FMNIST} \\
\midrule
NeuPSL~\cite{pryor2022neupsl}
& $0.88\pm0.02$
& $0.79\pm0.09$
& $0.65\pm0.12$
& $0.74\pm0.04$ \\

LTN-IND A~\cite{morra2023ltn}
& $0.83\pm0.18$
& $0.58\pm0.04$
& $0.83\pm0.09$
& $0.67\pm0.11$ \\

LTN-IND B~\cite{morra2023ltn}
& $0.84\pm0.14$
& $0.58\pm0.06$
& $0.85\pm0.11$
& $0.76\pm0.15$ \\

LTN-IND C~\cite{morra2023ltn}
& $\mathbf{0.94\pm0.10}$
& $0.65\pm0.14$
& $0.87\pm0.09$
& $0.83\pm0.11$ \\

\rowcolor{gray!20}
\textbf{Ours}
& $\mathbf{0.94\pm0.10}$
& $\mathbf{0.93\pm0.10}$
& $\mathbf{0.88\pm0.10}$
& $\mathbf{0.87\pm0.09}$ \\
\bottomrule
\end{tabular}
\end{table}

Table~\ref{tab:comparison} shows that the proposed method consistently matches or outperforms the baselines across all four domains. Critically, these gains arise even though the induced rule is logically equivalent to the one used by the LTN-IND variants~\cite{morra2023ltn}, so they cannot be attributed to differences in symbolic knowledge. 
We attribute the observed performance improvements over \cite{morra2023ltn} to two main architectural differences: (i)~our visual encoder produces high-dimensional \emph{embeddings} rather than digit predictions, and (ii) we performed a substantially broader hyperparameter search, resulting in a more robust visual encoder optimized for the low data regime of the benchmark. However, architectural differences in the encoder are minimal, while the grounding of the perceptual predicate \texttt{P\_same\_value} differs substantially: LTN-IND computes an exponential distance on logit predictions, whereas our visual encoder computes the same exponential distance function in a high-dimensional embedding space. On visually harder domains (e.g., EMNIST) embedding-distance grounding does not require correct classification and degrades far more gracefully. More broadly, the result highlights how the choice of grounding space can substantially affect the effectiveness of downstream logical verification, and the potential for co-design of logic formulation and grounding.

\subsection{Discussion}

The results demonstrate that combining VLM-based symbolic hypothesis generation with differentiable logical verification enables the discovery of interpretable visual reasoning rules from limited supervision. An important observation is that the framework does not necessarily recover the exact symbolic formulation used to define the benchmark constraint, but instead discovers alternative formulations that are logically equivalent. This indicates that the proposed approach performs a search over valid symbolic explanations rather than attempting to reproduce a predefined rule. The ability to converge to semantically equivalent representations despite syntactic variations highlights the benefit of coupling the generative capabilities of VLMs with a formal verification mechanism. 

The verification process plays a central role in constraining the space of possible symbolic hypotheses. While VLMs provide the flexibility required to propose candidate FOL expressions, they do not inherently guarantee logical correctness or consistency with the visual evidence. By dynamically translating generated rules into differentiable logical models, the D-LTN provides a semantic evaluation signal that guides the iterative refinement process. This interaction suggests that VLMs can be effectively employed as symbolic hypothesis generators when combined with mechanisms capable of evaluating and correcting their outputs. 

The analysis of the FMNIST domain highlights an important limitation of the current framework. Although the VLM generated a syntactically valid FOL expression, the resulting rule was semantically insufficient, as it encoded only a partial representation of the target Sudoku constraint (see Listing~\ref{list:rules_fail}). We hypothesize that this failure primarily arises from the VLM difficulties in establishing a reliable visual-to-symbolic correspondence in this domain. Compared with handwritten digit datasets, FMNIST contains visually more complex and semantically diverse categories, making it more difficult for the VLM to establish a reliable mapping between visual observations and symbolic predicates from only a few examples. As a result, although the induced rule was syntactically correct, its grounding was based on an inaccurate visual abstraction, leading to an incomplete representation of the underlying Sudoku constraint. Overall, the findings indicate that VLMs can support automatic symbolic knowledge acquisition when integrated with formal verification, reducing the need for manually specified rules. 

\begin{figure}[t]
\label{fig:rules_fail}
\begin{lstlisting}[caption={Invalid symbolic rule discovered by the VLM for the FMNIST domain within the 20 iterations. Although syntactically valid, the rule is semantically insufficient, as it does not completely encode the Sudoku constraints.}, label={list:rules_fail}]

RULE 4 (FMNIST):

    forall x1 forall x2
        (P_same_row(x1, x2) | P_same_col(x1, x2)) 
        implies !P_same_value(x1, x2)   
            
\end{lstlisting}
\end{figure}

\paragraph{Limitations and future works. } 

Despite promising results, the proposed framework has several limitations that open avenues for future work. First, the current approach assumes that the alphabet is fixed and provided a priori; integrating automatic symbol discovery---learning which concepts to ground as FOL predicates from raw visual data---would make the system more broadly applicable. Second, experiments are conducted in a single benchmark and in a single-rule scenario: extending the framework to handle multi-rule theories would better reflect the complexity of real-world visual reasoning tasks. Third, we do not explore transfer learning scenario, in which the perception module is pretrained, which may further simplify rule discovery \cite{daniele2024transfer}. Fourth, while the closed feedback loop bears structural similarity to reinforcement learning, no formal RL objective is optimized; adopting policy-gradient or actor-critic methods to train the VLM query strategy could make rule induction more sample-efficient. Finally, the framework is currently implemented as a standalone prototype; tighter integration with established NeSy libraries such as LTNtorch~\cite{badreddine2022ltn} or Uller~\cite{van2024uller} would lower the barrier to adoption and facilitate comparison with a broader range of baselines.

% ============================================================
\section{Conclusion}
\label{sec:conclusion}
% ============================================================

This paper presented a neuro-symbolic framework for automatic rule discovery from visual data. The proposed architecture integrates a Vision-Language Model (VLM) for First-Order Logic (FOL) rule induction with a Dynamic Logic Tensor Network (D-LTN) for differentiable rule verification within a closed iterative feedback loop. By combining neural perception, logical reasoning, and feedback-driven refinement, the system can generate, validate, and progressively improve candidate rules from only a small set of labeled visual examples.
Experiments on the ViSudo-PC benchmark showed that the framework successfully recovered valid Sudoku constraints in the MNIST, EMNIST, and KMNIST domains, consistently converging to logically equivalent formulations of the target rule despite syntactic variations. 
These findings demonstrate that VLMs can effectively support symbolic rule induction when coupled with a formal verification mechanism, reducing the need for manually specified knowledge. 
Overall, this work represents a step toward neuro-symbolic systems capable of autonomously acquiring symbolic knowledge from visual observations while preserving the interpretability and reasoning capabilities of logic-based models.

%\clearpage  % TODO FINAL: This \clearpage needs to be removed from both review and camera-ready versions.

\section*{Acknowledgements}
This study was partially supported by the ``WEBFARE'' (Nr. FISA2022-00908) project, funded by the Italian Ministry of University and Research under the FISA 2022 programme (D.D. No. 1405 of 13/09/2022, Fondo Italiano per le Scienze Applicate – FISA). This manuscript reflects only the authors’ views and opinions; the Ministry cannot be held responsible for them. 

\bibliographystyle{splncs04}
\bibliography{main}

\end{document}